\documentclass{article} % For LaTeX2e
\usepackage{iclr2026_conference,times}

\usepackage{amsmath,amsfonts,bm}

\def\eqref#1{equation~\ref{#1}}
\def\1{\bm{1}}

\DeclareMathAlphabet{\mathsfit}{\encodingdefault}{\sfdefault}{m}{sl}
\SetMathAlphabet{\mathsfit}{bold}{\encodingdefault}{\sfdefault}{bx}{n}

\usepackage{hyperref}
\usepackage{url}
\usepackage{graphicx}
\usepackage{subcaption}
\usepackage{booktabs}
\usepackage{multirow}
\usepackage{siunitx}
\usepackage{wrapfig} 

\usepackage{newtxmath}

\title{Right Regions, Wrong Labels: Semantic Label Flips in Segmentation under Correlation Shift}

\author{%
\begin{tabular}[t]{c}
\textbf{Akshit Achara}\hspace{1.5em}%
\textbf{Yovin Yahathugoda}\hspace{1.5em}%
\textbf{Nicholas Byrne}\hspace{1.5em}%
\textbf{Michela Antonelli}\\ \\ [0.2em]
\textbf{Esther Puyol Anton}\hspace{1.5em}%
\textbf{Alexander Hammers}\hspace{1.5em}%
\textbf{Andrew P.~King}\\ \\ [0.55em]
{\normalfont School of Biomedical Engineering \& Imaging Sciences, King's College London, UK}\\[0.25em]
\begin{tabular}[t]{c}
{\normalfont\mdseries\upshape\ttfamily\small
\{akshit.achara, yovin.yahathugoda, nicholas.byrne, michela.antonelli,}\\
{\normalfont\mdseries\upshape\ttfamily\small
esther.puyol\_anton, alexander.hammers, andrew.king\}@kcl.ac.uk}
\end{tabular}
\end{tabular}%
}

\iclrfinalcopy % Uncomment for camera-ready version, but NOT for submission.
\begin{document}

\maketitle

\begin{abstract}
The robustness of machine learning models can be compromised by spurious correlations between non-causal features in the input data and target labels. A common way to test for such correlations is to train on data where the label is strongly tied to some non-causal cue, then evaluate on examples where that tie no longer holds. This idea is well established for classification tasks, but for semantic segmentation the specific failure modes are not well understood. We show that a model may achieve reasonable overlap while assigning the wrong semantic label, swapping one plausible foreground class for another, even when object boundaries are largely correct. We focus on this semantic label-flip behaviour and quantify it with a simple diagnostic (Flip) that counts how often ground truth foreground pixels are assigned the wrong foreground identity while remaining predicted as foreground. In a setting where category and scene are correlated during training, increasing the correlation consistently widens the gap between common and rare test conditions and increases these within-object label swaps on counterfactual groups. Overall, our results motivate assessing segmentation robustness under distribution shift beyond overlap by decomposing foreground errors into correct pixels, flipped-identity pixels, and missed-to-background pixels. We also propose an entropy-based, ground truth label-free `flip-risk' score, which is computed from foreground identity uncertainty, and show that it can flag flip-prone cases at inference time. Code is available at \url{https://github.com/acharaakshit/label-flips}.
\end{abstract}

\section{Introduction}
\label{sec:intro}

Modern vision models can achieve high accuracy while leaning on `spurious' cues that are predictive of the target label in the training distribution but not causally tied to the task, a behaviour often described within a classification context as \emph{shortcut learning}~\citep{geirhos2020shortcut}. The spurious cues can be spatially structured, i.e. background context, imaging artefacts, and characteristic textures may correlate with target classes and provide an easier signal than object-centric evidence. When such correlations change at test time, performance can drop even if in-distribution results look strong.

Most spurious-correlation benchmarks and evaluation protocols were developed for classification, where a label is intentionally coupled with a nuisance attribute during training and then evaluated on balanced or counterfactual splits that break the association~\citep{sagawa2019distributionally}. Segmentation work has provided clear evidence that dense predictors can also exploit spurious correlations, for example those introduced by acquisition and preprocessing pipelines such as padding and cropping conventions or annotation markers~\citep{lin2024shortcut}. However, segmentation is often assessed with overlap-based metrics such as mIoU or Dice, which largely summarise geometric agreement. This emphasis can miss a failure mode specific to multi-class foreground segmentation: models may achieve reasonable foreground overlap yet assign the wrong semantic identity within that region. Conventional metrics will highlight this as a failure but obscure its true nature.

In this paper, we first demonstrate the existence of this phenomenon, which we refer to as semantic \emph{label flips} in segmentation. A label flip occurs when the prediction remains plausible as a foreground mask but swaps between foreground identities in response to a spurious cue. This differs from purely geometric failure modes where the mask collapses, drifts, or fragments. To isolate label flips under controlled distribution shift, we adapt the standard correlation-shift protocol to dense prediction: we vary the association strength between a binary label $Y$ and a binary attribute $\mathcal{A}$ during training, and evaluate on balanced splits that include all $(Y,\mathcal{A})$ combinations. This separates aligned conditions that dominate training from counterfactual conditions that break the correlation, enabling a direct test of whether semantic identity remains stable when cues change.

We conduct an analysis on two different datasets. First, we adapt \textsc{Waterbirds}~\citep{sagawa2019distributionally} to semantic segmentation, where background context provides a canonical shortcut that can affect semantic decisions without necessarily destroying localisation. We refer to this adapted dataset as \textsc{Waterbirds-seg}. Second, we study a real-mask setting built from confusable categories in \textsc{Coco}~\citep{lin2014microsoft}, where object identity (cat vs.\ dog) is correlated with coarse scene context (indoor vs.\ outdoor) derived from contextual labels. This is similar to contextual bias studied in natural-image shift benchmarks, where background context becomes spuriously tied to category (e.g., MetaShift-style evaluations)~\citep{liang2022metashift,lim2023biasadv}.  We refer to this dataset as \textsc{COCO-CD}. 

Across both datasets, increasing the training-time correlation consistently widens the performance gap between aligned and counterfactual groups and markedly increases within-foreground identity swaps, even when class-agnostic foreground overlap remains comparatively stable. Together, these results motivate assessing segmentation robustness under distribution shift beyond overlap by separating semantic flips from foreground deletion errors. Furthermore, we show that an entropy-based, ground truth label-free `flip-risk' score derived from foreground identity uncertainty can highlight flip-prone predictions for lightweight model monitoring.

\paragraph{Contributions.}
We make three contributions. First, we identify a segmentation failure mode under correlation shift in which models preserve plausible foreground extent but swap semantic identity between confusable foreground classes. Second, we adapt the standard correlation-shift protocol to dense prediction and instantiate it on two different datasets: a segmentation adaptation of \textbf{Waterbirds} and a real-mask \textbf{COCO} setting built from confusable categories with coarse indoor/outdoor context. Third, we complement group-wise overlap with diagnostics that separate semantic from geometric errors: for test-time evaluation with ground truth labels, we introduce Flip and a foreground-conditioned (correct/flip/miss) decomposition, and for deployment-time monitoring without ground truth labels, we propose an entropy-based flip-risk score from foreground identity uncertainty that stratifies flip-prone predictions and concentrates flips in a high-risk tail.

\section{Related Work}
\label{sec:related}

\subsection{Shortcut learning and correlation shift.}
Shortcut learning refers to models relying on predictive but non-causal cues that correlate with the target in the training distribution, but which can lead to brittle behaviour when those correlations change at test time~\citep{geirhos2020shortcut}. In image classification, this is commonly studied through correlation shift: training data is constructed so that the label is strongly associated with a spurious attribute, while evaluation uses balanced or counterfactual splits that break this association~\citep{sagawa2019distributionally}. Surveys and benchmarks have refined these protocols across vision tasks~\citep{ye2024spurious}. We use the same lens for dense prediction, but focus on a segmentation-specific consequence of cue shift: predictions can remain geometrically plausible while the semantic identity within the foreground becomes unstable.

\subsection{Spurious correlations and context dependence in semantic segmentation.}
Reliance on spurious correlations in segmentation has been documented when shortcut signals arise from the acquisition or annotation pipeline. For example, in medical image segmentation, \cite{lin2024shortcut} identified sources such as padding and preprocessing artefacts, cropping conventions, and annotation markers, and showed that segmentation models can exploit these signals despite their lack of anatomical meaning. More broadly, robustness work in dense prediction often studies failures under domain or style shift and reports degradation using overlap metrics such as Dice or mIoU; recent surveys summarise this landscape for semantic segmentation~\citep{rafi2024domain,schwonberg2025domain}. Overlap metrics such as Dice or mIoU capture how a model performs on segmentation, but they can blur the distinction between geometric failure and semantic confusion.

A complementary line of work studies context dependence in semantic segmentation directly. Related concerns have also been raised in adjacent object-centric learning work studying robustness to spurious background cues~\citep{rubinstein2025we}. In \cite{shetty2019not}, the authors quantify and control the influence of surrounding objects and scene context on segmentation predictions, and in \cite{hoyer2019grid}, the authors propose explanation maps designed for dense predictors that can detect and localise contextual biases. Related analyses also highlight biased context correlations as a central failure source in dense prediction, motivating methods that explicitly rectify context–label coupling~\citep{zhao2022rbc}. These results support the premise that context can steer segmentation predictions, but they do not isolate the specific failure mode we study: in multi-class foreground segmentation, a model may still outline a coherent object while assigning the wrong foreground label under a shift in correlated cues. Our work isolates this regime and measures it directly.

\subsection{Bias and subgroup disparities in segmentation.}
A complementary literature studies subgroup disparities in segmentation models, particularly in medical imaging, reporting differences across demographic groups (or \emph{biases}) for multiple anatomies and modalities~\citep{puyol2021fairness,puyol2022fairness,ioannou2022study,siddiqui2024fair,benvcevic2024understanding,alqarni2025investigation}. The race bias in cardiac MR segmentation \citep{puyol2021fairness,puyol2022fairness} was found to be at least partly due to differences in subcutaneous fat between races \citep{lee2025bias}, i.e. the model was exploiting these background features to guide its foreground segmentation predictions. Therefore, this is a similar phenomenon to that reported in \cite{lin2024shortcut} for other, non-demographic-related features. However, these bias studies summarise performance gaps using overlap-based metrics, which is appropriate for many clinical endpoints, but it leaves open a practical question about error structure under distribution shift. In segmentation, mistakes can stem from mislocalisation, from confusing foreground with background, or from assigning the wrong identity within a localised region. The label-flip behaviour we study falls into the last category and can therefore be easy to miss when reporting only aggregate overlap, even though it reflects a meaningful loss of semantic correctness under distribution shift.

\subsection{Uncertainty and selective prediction in segmentation.}
Uncertainty estimation for semantic segmentation has long been used to quantify prediction reliability and support risk-aware deployment, based on uncertainty from approximate Bayesian inference~\citep{kendall2017bayesian}.
Recent work has also studied calibration specifically for dense predictors and showed that segmentation networks can be overconfident, motivating post-hoc or training-time calibration methods~\citep{wang2023calibrating}.
Finally, selective prediction frameworks for semantic segmentation use confidence estimates to abstain or defer on uncertain predictions, and have been evaluated explicitly under distribution shift~\citep{borgesselective}.

\subsection{Context and co-occurrence bias.}
Contextual co-occurrence is a well-known driver of shortcut behaviour in vision. In weakly supervised semantic segmentation, pseudo-labelling pipelines often latch onto context or correlated objects, producing biased or incomplete masks; recent approaches mitigate this by separating co-occurring factors, enforcing consistency, or synthesising counterfactual object--background combinations~\citep{jo2023mars,zhang2025mitigating,kwon2024learning,yang2024separate}. Related contextual shift benchmarks construct many natural distributions defined by context metadata and have been used to study category-context entanglement, including cat/dog settings where indoor/outdoor context is spuriously correlated with the label~\citep{liang2022metashift,lim2023biasadv}. Co-occurrence has been studied in fully supervised semantic segmentation, where it can induce competing hypotheses and degrade dense labelling, motivating training schemes that explicitly account for co-occurrence~\citep{islam2023segmix}. Although motivated by weak supervision, this line of work reinforces a mechanism that carries over to fully supervised settings: context can steer semantic decisions even when localisation cues remain available. This is precisely the setting in which within-foreground identity swaps can arise under correlation shift.

% \subsection{Positioning.}
% Taken together, prior work has shown that segmentation models can exploit spurious cues and that performance can degrade under distribution shift, including in subgroup-dependent ways~\citep{puyol2021fairness,lin2024shortcut,benvcevic2024understanding}. Research has also demonstrated that segmentation predictions can be sensitive to contextual co-occurrence and that contextual biases can be quantified or diagnosed in dense predictors~\citep{shetty2019not,hoyer2019grid}. Our contribution is to make explicit a semantic failure mode that overlap metrics can obscure in multi-class foreground segmentation: under correlation shift, models may preserve foreground extent while swapping its identity. We therefore complement group-wise overlap with Flip and a foreground-conditioned error decomposition that separates correct pixels, flipped-identity pixels, and missed-to-background pixels, and we show that a ground truth label-free entropy-based flip-risk score derived from foreground identity uncertainty can help flag flip-prone predictions.

\section{Datasets and Experimental Setup}
\label{sec:benchmarks}

We study semantic label instability under controlled correlation shift in foreground segmentation with two confusable foreground identities and a shared background class. In both benchmarks, the label $Y$ denotes the foreground identity (landbird vs.\ waterbird; cat vs.\ dog) and the attribute $\mathcal{A}$ denotes a coarse context condition (land vs.\ water; indoor vs.\ outdoor) that can be spuriously correlated with $Y$ during training. This yields four groups $(Y,\mathcal{A}) \in \{0,1\}^2$ and enables evaluation on both aligned and counterfactual combinations.

\textbf{\textsc{Waterbirds-seg.}}
We adapt the standard \textsc{Waterbirds} construction to segmentation by using CUB~\citep{wah2011caltech} bird masks as pixel-level targets while retaining the original correlation mechanism between bird type and scene background. Here $Y$ is bird type and $\mathcal{A}$ is scene type (land vs.\ water). We consider two correlation regimes. For $\rho{=}0.5$, training is balanced across all four groups. For $\rho{=}0.95$, we construct a strongly biased training set with a $95\%$ association between $Y$ and $\mathcal{A}$ while matching the total number of training examples in the $\rho{=}0.5$ setting. Validation and test are balanced across groups in all cases. Full construction details are provided in Appendix~\ref{sec:appendix-data}.

\textbf{\textsc{COCO-CD.}}
To complement this synthetic-context benchmark, we build a real-mask setting from \textsc{COCO}. We merge the official splits and retain images containing either a cat or a dog, excluding images that contain both categories to avoid ambiguous supervision. We assign $\mathcal{A}$ (indoor vs.\ outdoor) from contextual annotations to obtain a coarse scene label, then subsample to induce a controllable association between object identity $Y$ and context $\mathcal{A}$. We defer the precise attribute assignment rule and resulting group statistics to Appendix~\ref{sec:appendix-data}.

\textbf{Training.}
We resize inputs to $512\times 512$; during training we apply standard geometric augmentations (random horizontal flip and random resized crops), and photometric jitter. We train U-Net~\citep{ronneberger2015u} models with two encoder families to avoid conclusions tied to a single backbone: a convolutional encoder (ResNet-50)~\citep{he2016deep} and a Transformer encoder (MiT-B2)~\citep{xie2021segformer}. Models are optimised with Adam~\citep{kingma2014adam} (learning rate $10^{-4}$, batch size $8$) for up to $200$ epochs with early stopping based on validation loss.

To reduce sensitivity to low-level nuisance variation while keeping comparisons controlled, we use the same augmentation pipeline in all experiments. This pipeline combines standard geometric and photometric transformations appropriate for dense prediction.
%and, in some settings, a background-targeted CutMix-style augmentation that perturbs contextual appearance while leaving the foreground supervision unchanged.
We compare four training variants that separate the role of the objective from context perturbations: cross-entropy (CE), Dice$+$cross-entropy (DCE)~\citep{milletari2016v}, Group DRO~\citep{sagawa2019distributionally} (GD) over the four $(Y,\mathcal{A})$ groups, and cross-entropy with a background-restricted CutMix-style augmentation (CM)~\citep{yun2019cutmix}. The CutMix variant modifies background regions only, thereby intervening on contextual cues without changing the foreground labels. All variants share the same base geometric and photometric augmentations for dense prediction, and we keep architectures and optimisation fixed across correlation regimes.
\subsection{Metrics and diagnostics}
\label{sec:metrics}

We report class-wise IoU and summarise overlap by macro mIoU over the two foreground classes.
To evaluate localisation independently of identity, we also report class-agnostic foreground IoU (Fg-IoU) by collapsing both foreground labels into a single foreground mask.

\paragraph{Quantifying semantic label flip.} To make semantic instability explicit, we measure within-foreground identity swaps.
We consider settings with background and two confusable foreground labels, $\mathcal{Y}=\{0,1,2\}$, where $0$ denotes background and $\{1,2\}$ are the foreground classes.
For each pixel $i$, let $y_i\in\mathcal{Y}$ be the ground truth label and $\hat{y}_i\in\mathcal{Y}$ the prediction.
We define the within-foreground flip rate as
\begin{equation}
\label{eq:flip}
\mathrm{Flip}(\hat{y},y)
=
\frac{
\sum_i \mathbb{I}[y_i \in \{1,2\}]\,\mathbb{I}[\hat{y}_i \in \{1,2\}]\,\mathbb{I}[\hat{y}_i \neq y_i]
}{
\sum_i \mathbb{I}[y_i \in \{1,2\}]
},
\end{equation}
where $\mathbb{I}[\cdot]$ is the indicator function.
This counts identity swaps on ground truth foreground pixels that remain predicted as foreground.
We compute Flip globally and per subgroup $(Y,\mathcal{A})$ on balanced evaluation splits and summarise robustness by contrasting aligned versus counterfactual conditions.

\paragraph{Foreground-conditioned error decomposition.}
Because Flip only counts pixels that remain predicted as foreground, we also decompose errors over ground truth foreground pixels to separate identity swaps from foreground deletion.
Let $\mathcal{F}=\{x:\,y(x)\in\{1,2\}\}$ be the set of ground truth foreground pixels with size $|\mathcal{F}|$.
We define
\begin{align}
\mathrm{FG\text{-}Corr} \;&=\; \frac{1}{|\mathcal{F}|}\sum_{x\in\mathcal{F}} \mathbb{I}\!\left[\hat y(x)=y(x)\right], \\
\mathrm{FG\text{-}Flip} \;&=\; \frac{1}{|\mathcal{F}|}\sum_{x\in\mathcal{F}} \mathbb{I}\!\left[\hat y(x)\in\{1,2\}\ \wedge\ \hat y(x)\neq y(x)\right], \\
\mathrm{FG\text{-}Miss} \;&=\; \frac{1}{|\mathcal{F}|}\sum_{x\in\mathcal{F}} \mathbb{I}\!\left[\hat y(x)=0\right],
\end{align}
so that $\mathrm{FG\text{-}Corr}+\mathrm{FG\text{-}Flip}+\mathrm{FG\text{-}Miss}=1$ by construction.
We report this breakdown to distinguish semantic identity errors (FG-Flip) from foreground deletion/localisation failures (FG-Miss).
Full tables are reported in Appendix~\ref{sec:appendix-fgdecomp}.

\paragraph{Ground truth label-free flip-risk score for monitoring.}
To flag flip-prone predictions without ground truth labels (e.g. at deployment-time), we compute a flip-risk score from model outputs.
For each pixel, we renormalise predicted probabilities over the two foreground classes,
$p^t_k(i)=p_k(i)/(p_1(i)+p_2(i))$ for $k\in\{1,2\}$,
and define pixel risk as the binary entropy
\begin{equation}
\label{eq:risk}
r(i) = -\sum_{k\in\{1,2\}} p^t_k(i)\log p^t_k(i).
\end{equation}
We obtain an image-level flip-risk by averaging $r(i)$ over predicted-foreground pixels (those with $\hat{y}_i\in\{1,2\}$).
To evaluate the flip-risk score, we stratify images into 10 equal-sized risk deciles (low to high) and report the mean per-image Flip within each decile.
Note that the ground truth labels are used only for the computation of Flip and the error decomposition; flip-risk is computed from predictions only and hence can be monitored at inference time without ground truth labels.
 
\section{Experiments}
\label{sec:experiments}

We study whether correlation shift can induce semantic label flips, meaning identity swaps within the foreground, even when the predicted foreground extent remains plausible.

Our experiments address three questions:
(i) how increasing spurious correlation widens aligned vs.\ counterfactual gaps and changes error structure beyond overlap metrics,
(ii) whether the resulting errors reflect semantic instability conditioned on context (identity swaps with largely preserved foreground extent), rather than foreground deletion or gross localisation failure,
and (iii) whether flip-prone predictions can be flagged at deployment-time from model outputs alone, using a ground truth label-free flip-risk score.

Across benchmarks, we train with correlation strength $\rho \in \{0.5,0.95\}$ and evaluate on balanced validation and test splits containing all four $(Y,\mathcal{A})$ groups.
%We use a U-Net architecture with two encoder families (ResNet-50 and MiT-B2).
Unless otherwise noted, we focus on $\rho{=}0.95$ as a stress test, and use $\rho{=}0.5$ as a balanced reference to isolate the effect of stronger spurious coupling.

We evaluate subgroup disparities using IoU/mIoU and isolate semantic instability using Flip, a foreground error decomposition, and a ground truth label-free flip-risk score (Section~\ref{sec:metrics}).
We first quantify aligned versus counterfactual IoU gaps, then visualise label flips qualitatively, probe the mechanism with masking interventions, and finally test whether the flip-risk can stratify flips for model monitoring purposes.

\begin{wrapfigure}{r}{0.6\textwidth}
\vspace{-0.25cm}
\centering
\resizebox{.6\textwidth}{!}{
\includegraphics[width=\linewidth]{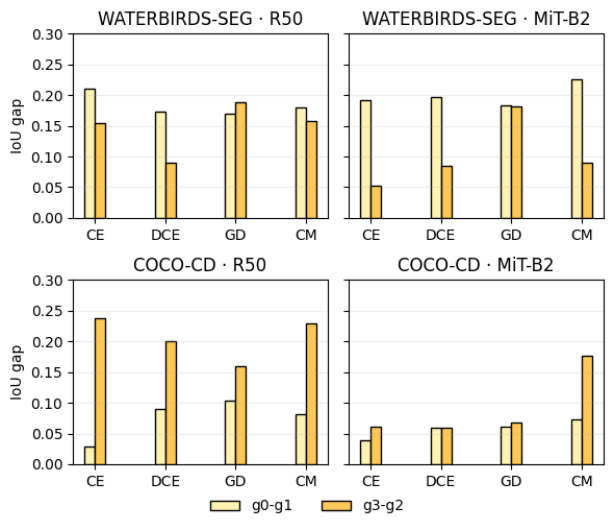}}
\caption{\textbf{Subgroup class IoU disparities under $\rho{=}0.95$.}
On the balanced evaluation split, each bar corresponds to a specific foreground class and reports the IoU gap between the aligned and counterfactual subgroup for that class. Here, $g0-g1$ denotes \emph{landbird+land - landbird+water} for \textsc{Waterbirds-Seg} and \emph{cat+indoor - cat+outdoor} for \textsc{COCO-CD}, while $g3-g2$ denotes \emph{waterbird+water - waterbird+land} for \textsc{Waterbirds-Seg} and \emph{dog+outdoor - dog+indoor} for \textsc{COCO-CD}.}

\label{fig:disparities}
\vspace{-0.25cm}
\end{wrapfigure}

\paragraph{Demonstrating semantic label flips under correlation shift.}
Under $\rho{=}0.95$, all objectives exhibit consistent aligned versus counterfactual disparities across both benchmarks and encoder families.
Overlap-based metrics capture this as reduced mIoU on counterfactual groups.
Figure~\ref{fig:disparities} makes this more granular by reporting, for each foreground class, the aligned--counterfactual IoU gap: per-class IoU is consistently higher on aligned groups and lower on counterfactual groups.
Importantly, these disparities can persist even when overall class-wise IoUs improve under alternative objectives.
Absolute class-wise IoUs are reported in Appendix~\ref{app:summary_stats} to show that the approaches might help with overall IoU while the disparities remain.
In contrast, when training is balanced at $\rho{=}0.5$, these aligned versus counterfactual IoU gaps are much smaller across settings (Appendix~\ref{app:additional_results}), isolating correlation strength as the driver of the observed subgroup disparities.  

To distinguish semantic identity swaps from foreground deletion, we also report the foreground-conditioned error decomposition over ground truth foreground pixels (Section~\ref{sec:metrics}). Unless otherwise stated, we compute FG-Corr/FG-Flip/FG-Miss globally over the balanced evaluation split, i.e., across all (Y,A) groups. This breakdown partitions pixels into correct identity (FG-Corr), flipped identity while remaining foreground (FG-Flip), and missed-to-background (FG-Miss), which sum to 1 by construction.
Table~\ref{tab:fgdecomp_waterbirds} shows that on \textsc{Waterbirds-seg}, increasing correlation from $\rho{=}0.5$ to $\rho{=}0.95$ primarily increases FG-Flip while FG-Miss remains comparatively small across objectives and encoders, supporting the claim that semantic instability can arise even when foreground extent is largely preserved.
We report the corresponding decomposition for \textsc{COCO-CD} in Appendix~\ref{sec:appendix-fgdecomp}.

\begin{table}[t]
\centering
\small
\setlength{\tabcolsep}{4.5pt}
\resizebox{.75\textwidth}{!}{
\begin{tabular}{ll ccc ccc}
\toprule
Model & Loss &
\multicolumn{3}{c}{$\rho=0.5$} &
\multicolumn{3}{c}{$\rho=0.95$} \\
\cmidrule(lr){3-5}\cmidrule(lr){6-8}
& &
FG-Corr & FG-Flip & FG-Miss &
FG-Corr & FG-Flip & FG-Miss \\
\midrule
\multirow{4}{*}{R50}
& CE  & 0.916 & 0.057 & 0.026 & 0.834 & 0.126 & 0.040 \\
& DCE & 0.918 & 0.053 & 0.029 & 0.850 & 0.117 & 0.033 \\
& GD  & 0.910 & 0.054 & 0.037 & 0.846 & 0.117 & 0.037 \\
& CM  & 0.915 & 0.051 & 0.034 & 0.838 & 0.117 & 0.045 \\
\midrule
\multirow{4}{*}{MiT-B2}
& CE  & 0.915 & 0.061 & 0.024 & 0.871 & 0.100 & 0.029 \\
& DCE & 0.942 & 0.036 & 0.022 & 0.868 & 0.100 & 0.031 \\
& GD  & 0.931 & 0.045 & 0.024 & 0.858 & 0.115 & 0.027 \\
& CM  & 0.935 & 0.039 & 0.026 & 0.872 & 0.100 & 0.027 \\
\bottomrule
\end{tabular}}
\caption{\textbf{\textsc{Waterbirds-seg} foreground error decomposition.}
Rates are computed over ground truth foreground pixels and satisfy FG-Corr + FG-Flip + FG-Miss = 1. Notably, increasing $\rho$ increases FG-Flip more than FG-Miss, indicating semantic instability despite preserved foreground extent.}
\label{tab:fgdecomp_waterbirds}
\end{table}

\begin{wrapfigure}{r}{0.5\textwidth}
% \vspace{-0.75cm}
\centering
\resizebox{.5\textwidth}{!}{
\includegraphics[width=\linewidth]{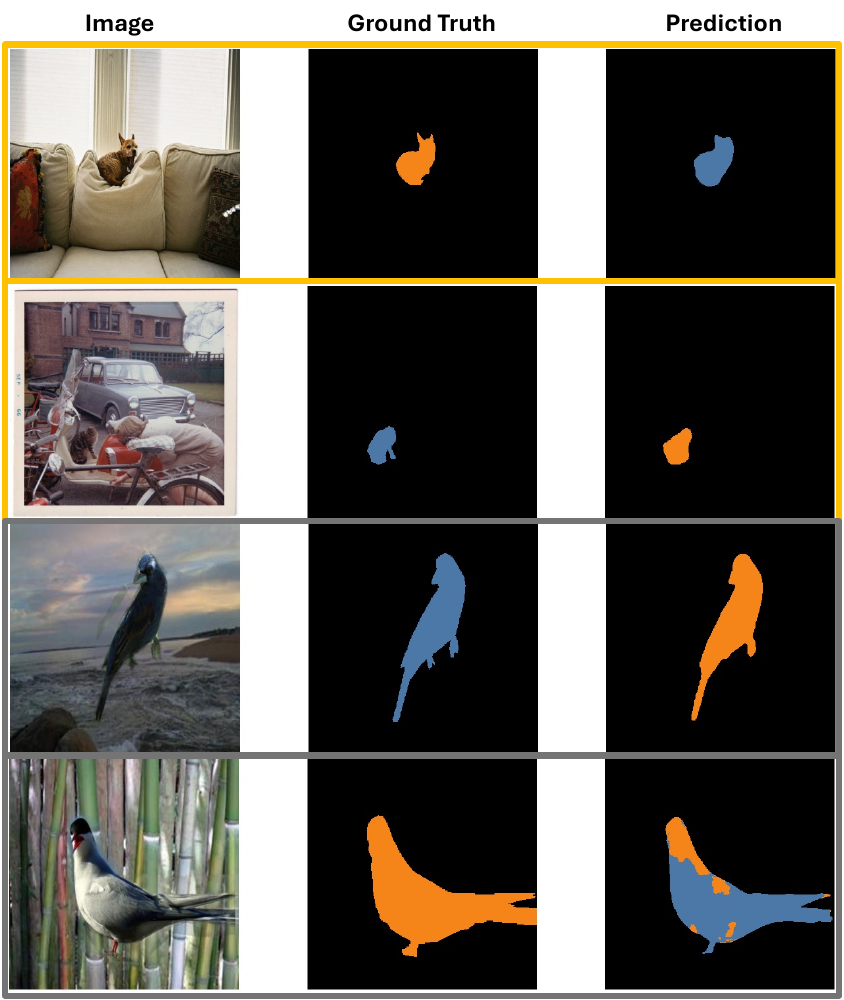}}
\caption{\textbf{Qualitative flips.} Extent is preserved while identity swaps under a counterfactual context.}
\label{fig:qual}
% \vspace{-0.5cm}
\end{wrapfigure}

\paragraph{Qualitative illustration of semantic label flips.}
Figure~\ref{fig:qual} visualises the failure mode behind the subgroup gaps by showing cases where the predicted foreground region remains largely correct, but the assigned identity changes with context.
On \textsc{COCO-CD}, counterfactual context can drive systematic swaps, such as an indoor dog being segmented as a cat or an outdoor cat being segmented as a dog.
On \textsc{Waterbirds-seg}, we observe partial flips within the same object, where parts of a groundbird are labelled as waterbird and parts of a waterbird are labelled as groundbird.
These examples highlight that correlation shift can induce semantic instability within the foreground even when the overall geometry is mostly intact. Appendix~\ref{app:oracle} further probes the mechanism using mask-controlled mean-fill interventions (Oracle-FG/Oracle-BG), comparing behaviour when background is removed versus when foreground is removed at inference time.

\paragraph{Monitoring flips via an inference time flip-risk score.}
Finally, we ask whether flip-prone predictions can be flagged from model outputs at inference time. We sort images by a flip-risk score (Section~\ref{sec:metrics}), split them into 10 equal-sized deciles (low to high), and report the mean per-image Flip within each decile.

\begin{figure}[t!]
% \vspace{-1cm}
\centering
\includegraphics[width=\linewidth]{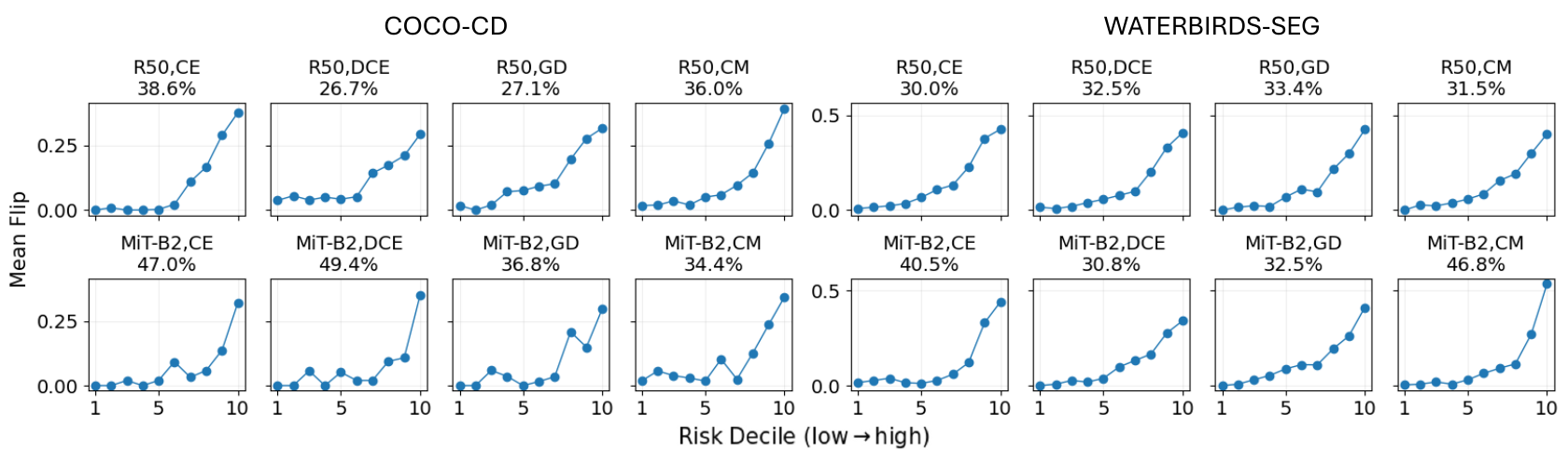}
\caption{\textbf{Flip-risk deciles ($\rho=0.95$).} Images are binned by predicted-foreground entropy risk; Flip concentrates in the highest-risk tail. Titles report top-decile flip share. R50 is short for ResNet50 here.}
\label{fig:risk_deciles}
\end{figure}

Figure~\ref{fig:risk_deciles} shows that this ground truth label-free score consistently stratifies semantic instability across both benchmarks, both encoder families, and all objectives considered.
Flip remains near zero in low-risk deciles and rises sharply in the highest-risk tail. Balanced training ($\rho{=}0.5$; Appendix~\ref{app:additional_results}) exhibits lower Flip overall, with flips largely confined to the highest-risk deciles.
In contrast, under $\rho{=}0.95$ the mean Flip is higher and non-trivial Flip appears already at earlier deciles, indicating that stronger spurious coupling broadens semantic instability beyond the extreme tail. Across objectives, mitigations often reduce Flip in the low to mid risk deciles, indicating improved semantic stability for the majority of predictions.
However, the highest-risk tail remains challenging: in most settings, high-risk deciles still exhibit substantial Flip even with Group DRO or CutMix, suggesting that these interventions reduce average instability but do not fully eliminate the most flip-prone cases. Appendix~\ref{app:mask_conditioned_risk} shows that the decile stratification is stable to how risk is aggregated: computing risk over predicted-foreground pixels (deployable) or ground-truth foreground pixels (analysis) yields the same qualitative trend, and increasing $\rho$ shifts flips into lower-risk deciles.

\section{Discussion}
\label{sec:discussion}

Our results highlight a segmentation-specific failure mode under correlation shift that overlap metrics can understate.
When identity and context are spuriously coupled during training, models can produce masks with plausible geometry while semantics become context contingent, swapping foreground identity under counterfactual context.
This decoupling explains why class-agnostic localisation can remain comparatively stable while within-foreground flips increase on minority groups.

The strong-correlation regime ($\rho{=}0.95$) serves as a stress test that isolates reliance on context as a proxy for identity.
With balanced evaluation and increased training-time association, we observe a shift in error composition: overlap aggregates missed extent, boundary misalignment, and identity errors, whereas Flip targets semantic instability within the foreground.

The behaviour across objectives provides additional nuance.
Dice+CE optimises per-class overlap and penalises identity swaps, but strong overlap alone does not guarantee semantic stability under correlation shift.
Group DRO improves worst-group performance by re-weighting higher-loss groups, yet flips can persist, indicating that worst-group optimisation alone does not guarantee disentanglement of identity from context.
Overall, improved overlap and stable semantics are related but not equivalent targets.

The flip-risk decile analysis suggests a practical handle for monitoring this failure mode.
A ground truth label-free uncertainty score derived from the two-way foreground identity distribution stratifies flips in both regimes, with flips appearing earlier and at higher levels under $\rho{=}0.95$.
This does not remedy spurious feature reliance, but it provides a lightweight triage signal for selective review or deferral on high-risk predictions.

There are clear limitations.
We study the smallest multi-class setting that permits semantic flipping (two confusable foreground classes with a binary context attribute) and do not evaluate a domain-specific end use case.
We intentionally use simplified settings to make the failure mode observable and comparable under controlled correlation shift; future work should test how it manifests in more realistic segmentation pipelines and what downstream impact it has.

The main takeaway is methodological: robustness assessment for segmentation under distribution shift should distinguish localisation collapse from semantic identity drift within correctly localised regions.
Flip makes the semantic component explicit; the foreground-conditioned decomposition (FG-Corr/FG-Flip/FG-Miss) separates identity swaps from foreground deletion; and the ground truth label-free flip-risk score enables deployment-time monitoring by flagging predictions most likely to contain semantic instability.

\section{Conclusion}
\label{sec:conclusion}

We have revisited spurious correlations in semantic segmentation under a controlled correlation-shift protocol and identified a segmentation-specific failure mode that overlap metrics can understate.
When identity and context are spuriously coupled, models may preserve a plausible foreground extent while swapping semantic identity within that extent, yielding masks that appear geometrically correct but are semantically incorrect.
To make this behaviour explicit, we report a within-foreground Flip statistic that measures identity swaps inside the true foreground while excluding background confusions, alongside standard overlap metrics, class-agnostic foreground IoU, and an FG-Corr/FG-Flip/FG-Miss decomposition that separates identity swaps from missed-to-background errors.

Across \textsc{Waterbirds-seg} and \textsc{COCO-CD} with real masks, increasing training-time correlation consistently amplified flips on counterfactual groups, often more sharply than changes in geometry-focused overlap.
We further observed that common training objectives and worst-group reweighing can change the balance between localisation quality and semantic stability, reinforcing that improved overlap does not necessarily imply robust identity under shift.
Finally, an entropy-based flip-risk score derived from the model’s two-way foreground distribution concentrated flips in a high-risk tail, suggesting a lightweight mechanism for monitoring or triaging predictions when semantic identity is critical.

Overall, these results motivate a practical recommendation for evaluating dense predictors under distribution shift: assess robustness not only through geometric overlap, but also through semantic stability within correctly localised regions. We hope these diagnostics encourage more fine-grained reporting of semantic failures in segmentation under spurious correlations.

\section{Acknowledgments}
This research was supported by the UK Engineering and Physical Sciences Research Council (EPSRC) [Grant reference number EP/Y035216/1] Centre for Doctoral Training in Data-Driven Health (DRIVE-Health) at King’s College
London.

\bibliography{iclr2026_conference}
\bibliographystyle{iclr2026_conference}

\appendix
% \section{Appendix}

\clearpage
\section{Dataset Construction Details}
\label{sec:appendix-data}

\paragraph{\textsc{Waterbirds-seg}.}
We use CUB bird masks for dense supervision and follow the \textsc{Waterbirds} group construction with $\rho\in\{0.5,0.95\}$, matching train size across regimes and keeping validation/test balanced across groups. Table~\ref{tab:wb-splits} lists the resulting group counts for train/validation/test under both correlation regimes.

\begin{wraptable}{r}{0.5\textwidth}
\vspace{-0.5cm}
\centering
\caption{\textbf{\textsc{Waterbirds-seg} split sizes.} Group counts for each split and correlation regime.}
\label{tab:wb-splits}
\resizebox{.5\textwidth}{!}{
\begin{tabular}{lcccc}
\toprule
Split & Landbird+Land & Landbird+Water & Waterbird+Land & Waterbird+Water \\
\midrule
Train ($\rho{=}0.5$)  & 615 & 615 & 615 & 615 \\
Train ($\rho{=}0.95$) & 1167 & 62 & 62 & 1167 \\
Val        & 74 & 74 & 74 & 74 \\
Test       & 642 & 642 & 642 & 642 \\
\bottomrule
\end{tabular}}
\end{wraptable}

\paragraph{\textsc{COCO-CD}.}
We construct \textsc{COCO-CD} from exclusive cat/dog images and derive an indoor/outdoor scene attribute from \textsc{COCO-Stuff} via the evidence aggregation procedure below, retaining only images with a clear indoor/outdoor context, then form $\rho\in{0.5,0.95}$ training regimes with balanced evaluation. Table~\ref{tab:coco-splits} lists the resulting group counts for train/validation/test under both correlation regimes.

\begin{wraptable}{r}{0.5\textwidth}
\centering
\caption{\textbf{\textsc{COCO-CD} split sizes.} Group counts per split and correlation regime.}
\label{tab:coco-splits}
\resizebox{.5\textwidth}{!}{
\begin{tabular}{lcccc}
\toprule
Split & Cat+Indoor & Cat+Outdoor & Dog+Indoor & Dog+Outdoor \\
\midrule
Train ($\rho{=}0.5$)  & 417 & 417 & 417 & 417 \\
Train ($\rho{=}0.95$) & 792 & 42 & 42 & 792 \\
Val        & 50 & 50 & 50 & 50 \\
Test       & 125 & 125 & 125 & 125 \\
\bottomrule
\end{tabular}}
\end{wraptable}

\paragraph{Context labelling via evidence aggregation.}
Each \textsc{COCO-Stuff} category is mapped to one of two values:
\texttt{indoor} if it appears in a curated indoor list and, \texttt{outdoor} if it appears in a curated outdoor list.
Given an image, we aggregate contextual evidence by summing the annotated pixel areas assigned to indoor-mapped and outdoor-mapped categories.

\begin{wraptable}{r}{0.5\textwidth}
\centering
\caption{\textbf{\textsc{COCO-CD} context coverage.} Proportion of retained (exclusive) images assigned to each context label by the evidence procedure.}
\label{tab:coco-context-coverage}
\resizebox{.45\textwidth}{!}{
\begin{tabular}{lccc}
\toprule
Subset & Indoor (\%) & Outdoor (\%) & Unknown/Excluded (\%) \\
\midrule
All retained & 38.13 & 46.40 & 15.47 \\
Cats only & 71.10 & 4.80 & 24.10 \\
Dogs only & 5.16 & 88.01 & 6.83 \\
\bottomrule
\end{tabular}}
\end{wraptable}

We assign $\mathcal{A}{=}\texttt{indoor}$ (respectively \texttt{outdoor}) when (i) the total contextual evidence covers a non-trivial portion of the image and (ii) indoor evidence (respectively outdoor evidence) dominates by a fixed ratio threshold.
Images with insufficient evidence are excluded from context-controlled splits. We additionally require every image to have at least $1\%$ foreground in the image post resizing.
This yields a coarse scene attribute that is independent of foreground mask construction while remaining visually meaningful for auditing under shift.
Table~\ref{tab:coco-context-coverage} reports the resulting coverage of the indoor/outdoor assignment on the retained exclusive cat/dog pool, including the fraction labelled \texttt{unknown} and excluded.

\paragraph{Manual validation of context labels.}
To assess the reliability of the derived indoor/outdoor attribute $\mathcal{A}$, we manually audited a stratified sample of $200$ validation images ($50$ per subgroup: Cat+Indoor/Outdoor and Dog+Indoor/Outdoor).
Each image was labelled by visual inspection as \{\texttt{indoor}, \texttt{outdoor}, \texttt{ambiguous}\} based on global scene context.
Excluding ambiguous scenes (1\%), the automatic context labels agree with human judgments in 99\% of images (196/198), with $\ge$98\% agreement in each subgroup (see Table~\ref{tab:cococd_context_audit}).

\paragraph{Correlation regimes and balanced evaluation.}
Using $Y$ and $\mathcal{A}$, we form the four groups $(Y,\mathcal{A}) \in \{\texttt{cat},\texttt{dog}\} \times \{\texttt{indoor},\texttt{outdoor}\}$.
For $\rho{=}0.5$, the training split is balanced across these four groups by capping to the smallest subgroup.
For $\rho{=}0.95$, we enforce a $95\%$ association between $Y$ and $\mathcal{A}$ while matching the total number of training examples to the $\rho{=}0.5$ setting.
Validation and test splits are balanced across all four groups in both regimes to ensure aligned and counterfactual evaluation.

\begin{wraptable}{r}{0.5\textwidth}
\centering
\caption{\textbf{\textsc{COCO-CD} context audit.} Manual validation of indoor/outdoor labels on 200 validation images (50 per subgroup). Agreement excludes \texttt{ambiguous} scenes.}
\label{tab:cococd_context_audit}
\resizebox{.5\textwidth}{!}{
\begin{tabular}{lrrrr}
\toprule
Subgroup & $N$ & Ambig. (\%) & Correct / Non-ambig & Agree. (\%) $\uparrow$ \\
\midrule
Cat+Indoor & 50 & 2.0 & 48 / 49 & 98.0 \\
Cat+Outdoor & 50 & 2.0 & 48 / 49 & 98.0 \\
Dog+Indoor & 50 & 0.0 & 50 / 50 & 100.0 \\
Dog+Outdoor & 50 & 0.0 & 50 / 50 & 100.0 \\
\midrule
Overall & 200 & 1.0 & 196 / 198 & 99.0 \\
\bottomrule
\end{tabular}}
\end{wraptable}

\section{Additional Results for Balanced Training ($\rho{=}0.5$)}
\label{app:additional_results}

This appendix reports results for the balanced training regime $\rho{=}0.5$, which serves as a reference setting where $Y$ and $\mathcal{A}$ are not spuriously coupled in the training distribution.

Because validation and test splits are balanced across all $(Y,\mathcal{A})$ groups in both regimes, differences between $\rho{=}0.5$ and $\rho{=}0.95$ reflect the strength of the training-time association rather than changes in evaluation composition.
Relative to $\rho{=}0.95$, balanced training reduces reliance on context as a proxy for identity, and within-foreground flips are typically smaller and less concentrated on counterfactual groups.

\begin{wrapfigure}{r}{0.6\textwidth}
% \vspace{-1cm}
\centering
\resizebox{.5\textwidth}{!}{
\includegraphics[width=\linewidth]{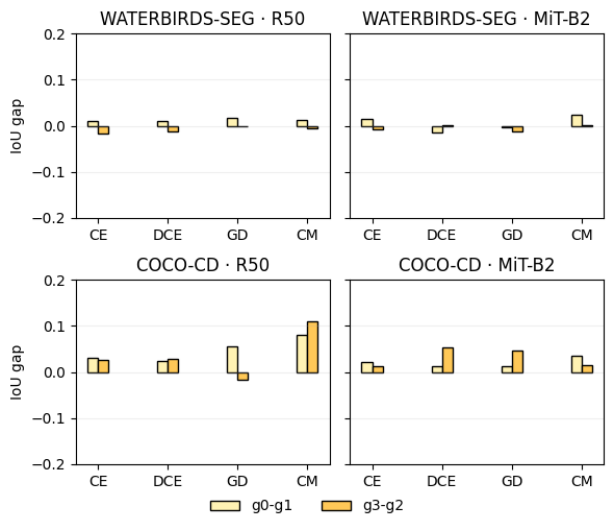}
}
\caption{\textbf{Balanced training ($\rho{=}0.5$): subgroup results.} Disparities are less pronounced than under $\rho{=}0.95$.}
\label{fig:disparities_rho05}
% \vspace{-0.25cm}
\end{wrapfigure}

Figure~\ref{fig:disparities_rho05} summarises subgroup performance and flip behaviour for models trained with $\rho{=}0.5$ and evaluated on the balanced test set.
Aligned versus counterfactual gaps can still arise due to residual dataset structure and finite-sample effects, but the overall level of semantic instability is lower than in the strongly correlated regime.

\begin{figure}[t]
\centering
\includegraphics[width=\linewidth]{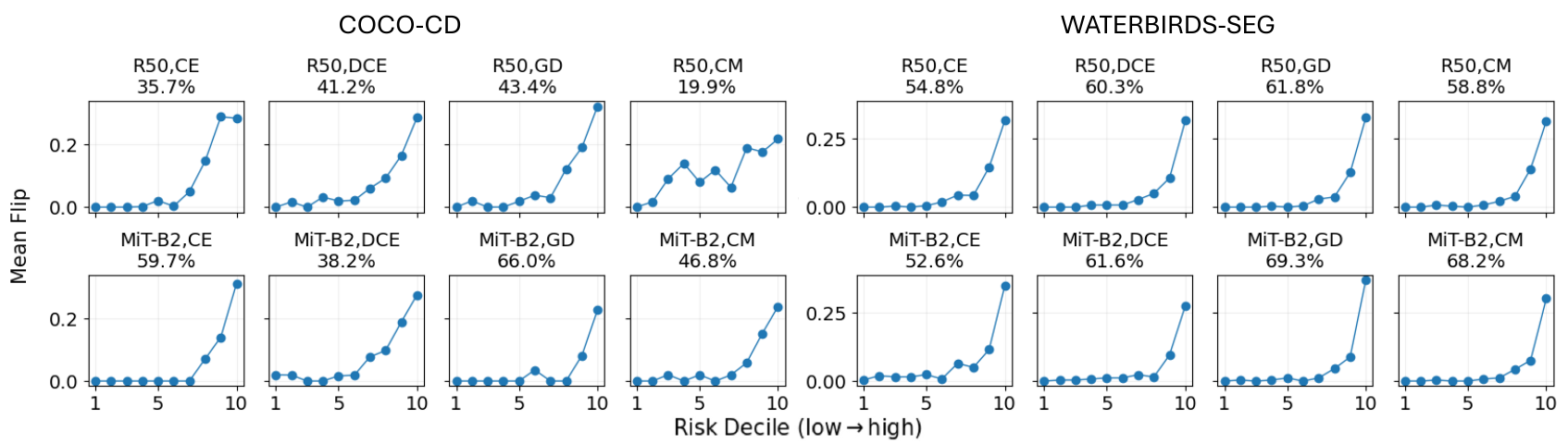}
\caption{\textbf{Balanced training ($\rho{=}0.5$): risk deciles.} Flip stratification persists but is weaker than under $\rho{=}0.95$.}
\label{fig:risk_deciles_rho05}
\end{figure}

We also report the flip-risk decile analysis under balanced training.
As in the main paper, we compute the entropy-based flip-risk score from the model's normalised probabilities over the two foreground identities and average it over predicted-foreground pixels.
We then sort images into equal-sized flip-risk deciles and report the mean per-image Flip within each decile.
Figure~\ref{fig:risk_deciles_rho05} shows that flip-risk still stratifies flips, but almost all the flips are concentrated in the highest-risk deciles. However, the Flips are lower than the corresponding rates for $\rho{=}0.95$.

\section{Masking Interventions}
\label{app:oracle}

To probe whether flips arise solely from background evidence at inference, or whether stronger correlation also induces foreground identity confusion, we apply two masking interventions using the ground-truth mask at evaluation.
In \textbf{Oracle-FG}, we preserve true foreground appearance and replace background pixels with a neutral mean-pixel fill.
In \textbf{Oracle-BG}, we preserve background context and replace foreground pixels with the same fill, so that predictions must rely on context alone.
We compare models trained with $\rho\in\{0.5,0.95\}$ and report $\Delta=(0.95)-(0.5)$.

\begin{wrapfigure}{r}{0.5\textwidth}
\vspace{-0.5cm}
\centering
\includegraphics[width=\linewidth]{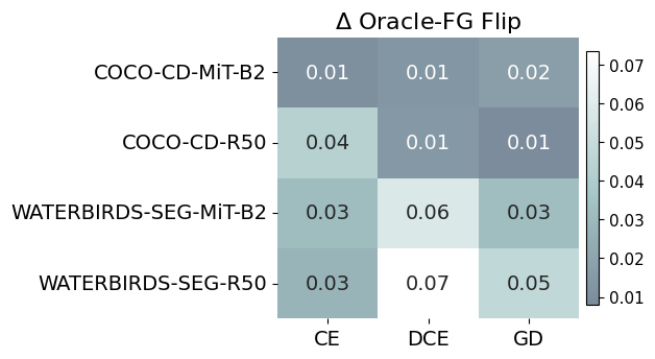}
\caption{\textbf{Oracle interventions.} $\Delta$Flip$(0.95{-}0.5)$ under mean-fill masking. Oracle-BG increases most with $\rho$ (background sufficiency); Oracle-FG increases more modestly, strongest on \textsc{Waterbirds-seg}.}
\label{fig:imprinting}
\end{wrapfigure}

Figure~\ref{fig:imprinting} reports $\Delta$ \textbf{Oracle-FG} Flip (background removed), isolating identity swaps when only foreground appearance is available.
Across settings, stronger correlation yields non-negative changes in Oracle-FG Flip, indicating that identity confusion can increase even when background evidence is unavailable at inference time.
We use ground-truth masks throughout to avoid conflating this mechanism probe with mask prediction errors.

We also ran the complementary \textbf{Oracle-BG} probe to test context-only behaviour.
On \textsc{Waterbirds-seg}, Oracle-BG retains high binary foreground IoU ($\gtrsim0.80$ under $\rho{=}0.95$), indicating that background context alone can localise the foreground region and making background sufficiency interpretable in this benchmark.
On \textsc{COCO-CD}, Oracle-BG can collapse to predicting background (low binary foreground IoU), so Flip under Oracle-BG may be mechanically suppressed; we therefore do not emphasise Oracle-BG quantitatively for \textsc{COCO-CD} and focus on Oracle-FG as the cleaner diagnostic of semantic instability under correlation shift.

Note that we omit CutMix from this analysis because it explicitly perturbs background appearance during training, which interacts with these evaluation-time masking probes and makes Oracle-BG effects less directly comparable.
A complementary representation analysis supports the same picture: as $\rho$ increases, the spurious attribute becomes more linearly accessible from pooled foreground encoder representations, with a stronger effect on \textsc{Waterbirds-seg} and a weaker but directionally consistent trend on \textsc{COCO-CD} (Appendix~\ref{app:pca-probe}).

\section{Attribute Accessibility in Foreground Features}
\label{app:pca-probe}

To complement the oracle masking interventions in Fig.~\ref{fig:imprinting}, we analyse whether the spurious attribute $\mathcal{A}$ becomes more linearly accessible in learned representations under strong correlation.

For each trained model and benchmark, we extract encoder feature maps on a balanced evaluation split and form per-image pooled representations by averaging features over ground truth foreground pixels and ground truth background pixels, yielding $z_{\mathrm{FG}}$ and $z_{\mathrm{BG}}$.
We fit PCA on a probe-train split (disjoint from probe-test), project $z_{\mathrm{FG}}$ or $z_{\mathrm{BG}}$ onto the top $k$ principal components, and train a logistic regression probe to predict $\mathcal{A}$.
We report test AUC as a function of $k$ for each training objective.

\begin{figure}[t!]
\centering
\includegraphics[width=\linewidth]{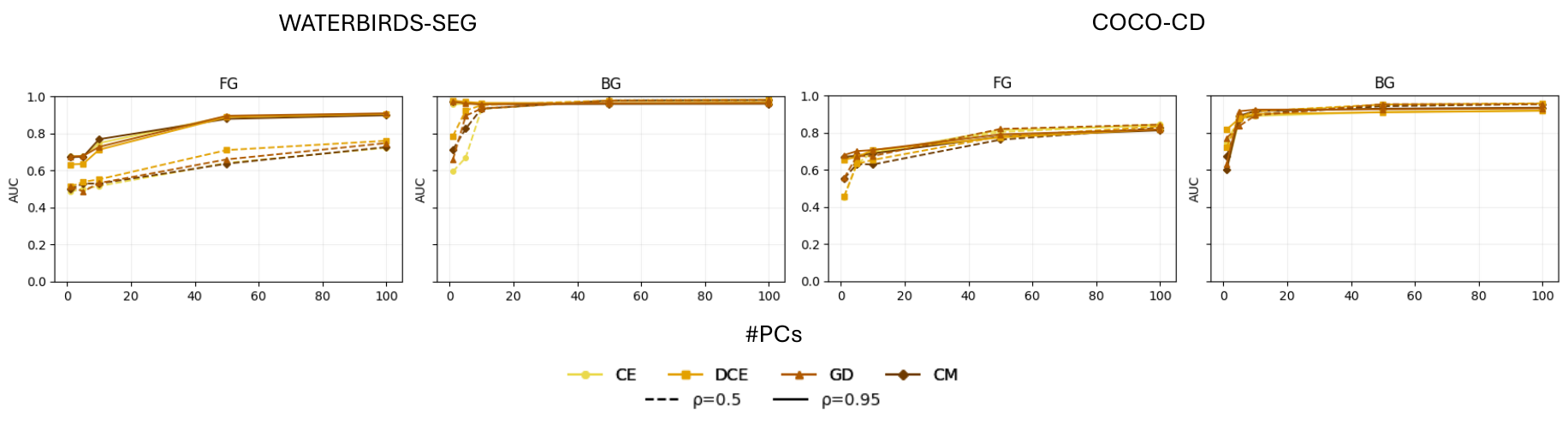}
\caption{\textbf{Attribute accessibility ($\rho{=}0.95$).} $\mathcal{A}$ is predictable from few PCs of $z_{\mathrm{FG}}$, while $z_{\mathrm{BG}}$ remains predictive. The results are computed using ResNet50.}
\label{fig:pca_probe}
\vspace{-0.3cm}
\end{figure}

Figure~\ref{fig:pca_probe} shows that, across both datasets, $\mathcal{A}$ is readily predictable from $z_{\mathrm{BG}}$, reflecting that background context naturally contains attribute information.
More notably, under $\rho{=}0.95$ the attribute becomes predictable from a few components of the foreground representation $z_{\mathrm{FG}}$, consistent with the Oracle-FG findings.

\section{Foreground Error Decomposition}
\label{sec:appendix-fgdecomp}

Our main analysis reports Flip, which isolates within-foreground identity swaps, but Flip does not count errors where ground truth foreground is predicted as background.
To separate these behaviours, we also report a foreground-conditioned decomposition over ground truth foreground pixels that partitions errors into correct identity (FG-Corr), flipped identity while remaining foreground (FG-Flip), and missed-to-background (FG-Miss).

\paragraph{Setup.}
We use the foreground-conditioned decomposition defined in Section~\ref{sec:metrics} and report FG-Corr/FG-Flip/FG-Miss across correlation regimes, objectives, and encoders.

\paragraph{Results on \textsc{COCO-CD}.}
The main paper reports this decomposition for \textsc{Waterbirds-seg} (Table~\ref{tab:fgdecomp_waterbirds}).
Here we provide the corresponding results on \textsc{COCO-CD} in Table~\ref{tab:fgdecomp_cococd}.
On \textsc{COCO-CD}, missed-to-background can be a substantial error mode for weaker models, so the decomposition clarifies whether performance losses stem from identity confusion (FG-Flip) or foreground deletion (FG-Miss).
ResNet-50 exhibits non-trivial FG-Miss across regimes (and an increase under $\rho{=}0.95$), alongside measurable FG-Flip.
In contrast, MiT-B2 reduces both FG-Flip and FG-Miss, indicating that a stronger backbone mitigates both identity swaps and foreground deletion in this real-mask setting.

\begin{table}[t]
\centering
\small
\setlength{\tabcolsep}{4.5pt}
\resizebox{.75\textwidth}{!}{
\begin{tabular}{ll ccc ccc}
\toprule
Model & Loss &
\multicolumn{3}{c}{$\rho=0.5$} &
\multicolumn{3}{c}{$\rho=0.95$} \\
\cmidrule(lr){3-5}\cmidrule(lr){6-8}
& &
FG-Corr & FG-Flip & FG-Miss &
FG-Corr & FG-Flip & FG-Miss \\
\midrule
\multirow{4}{*}{R50}
& CE  & 0.806 & 0.084 & 0.110 & 0.774 & 0.116 & 0.109 \\
& DCE & 0.814 & 0.075 & 0.111 & 0.767 & 0.101 & 0.132 \\
& GD  & 0.826 & 0.074 & 0.100 & 0.790 & 0.099 & 0.110 \\
& CM  & 0.700 & 0.149 & 0.151 & 0.779 & 0.111 & 0.110 \\
\midrule
\multirow{4}{*}{MiT-B2}
& CE  & 0.901 & 0.025 & 0.074 & 0.883 & 0.033 & 0.083 \\
& DCE & 0.877 & 0.053 & 0.070 & 0.869 & 0.045 & 0.086 \\
& GD  & 0.891 & 0.030 & 0.079 & 0.864 & 0.053 & 0.083 \\
& CM  & 0.883 & 0.030 & 0.087 & 0.838 & 0.077 & 0.085 \\
\bottomrule
\end{tabular}}
\caption{\textbf{\textsc{COCO-CD} foreground error decomposition.}
Rates are computed over ground truth foreground pixels and satisfy FG-Corr + FG-Flip + FG-Miss = 1.}
\label{tab:fgdecomp_cococd}
\end{table}

\section{Absolute Class-wise IoU}
\label{app:summary_stats}

For completeness, in Table~\ref{tab:iou_wrap}, we report absolute class-wise IoU values averaged over the balanced evaluation split.
These values provide context on overall performance across objectives, but our main conclusions focus on subgroup disparities and error composition under correlation shift (Fig.~\ref{fig:disparities}, Table~\ref{tab:fgdecomp_waterbirds}, and Appendix~\ref{sec:appendix-fgdecomp}).

\begin{table}[t!]
\centering
\caption{\textbf{Absolute class-wise IoU.} Each cell reports IoU$(C_1)$/IoU$(C_2)$, where $C_1$ and $C_2$ are the two foreground classes (Landbird/Waterbird for \textsc{Waterbirds-seg}; Cat/Dog for \textsc{COCO-CD}). The results are computed for $\rho=0.95$.}
\label{tab:iou_wrap}
\resizebox{0.75\linewidth}{!}{
\begin{tabular}{llcccc}
\toprule
Dataset & Model & CE & Dice+CE & GroupDRO & CutMix \\
\midrule
\multirow{2}{*}{\textsc{Waterbirds-seg}}
& R50 & 0.70/0.74 & 0.71/0.76 & 0.73/0.75 & 0.72/0.75 \\
& MiT-B2    & 0.74/0.78 & 0.75/0.78 & 0.74/0.76 & 0.75/0.79 \\
\midrule
\multirow{2}{*}{\textsc{COCO-CD}}
& R50 & 0.68/0.56 & 0.70/0.58 & 0.72/0.62 & 0.70/0.58 \\
& MiT-B2    & 0.82/0.75 & 0.81/0.74 & 0.79/0.74 & 0.77/0.69 \\
\bottomrule
\end{tabular}}
\end{table}

\section{Mask-conditioned risk analysis}
\label{app:mask_conditioned_risk}

The flip-risk score is computed from model outputs, but it aggregates uncertainty over a set of pixels.
In the main paper we average binary entropy over predicted-foreground pixels, which is deployment-friendly because it requires no ground truth.
Here we evaluate whether the stratification behaviour depends on this choice by recomputing risk over the ground-truth foreground pixels.

Concretely, we define two image-level risks:
(i) $\mathrm{Risk}_{\text{pred-FG}}$, the mean entropy over pixels predicted as foreground, and
(ii) $\mathrm{Risk}_{\text{GT-FG}}$, the mean entropy over ground-truth foreground pixels.
We then bin images into deciles under each risk and measure within-foreground Flip in each decile (Flip itself is label-dependent by definition).
Across datasets and objectives, $\mathrm{Risk}_{\text{GT-FG}}$ yields the same qualitative stratification as $\mathrm{Risk}_{\text{pred-FG}}$:
higher-risk deciles contain substantially more within-foreground identity swaps, and stronger correlation shifts flips into lower-risk deciles.
This indicates that the risk score is capturing foreground identity uncertainty rather than being driven primarily by errors in predicted-foreground support.

\begin{figure}[t!]
\centering
\includegraphics[width=\linewidth]{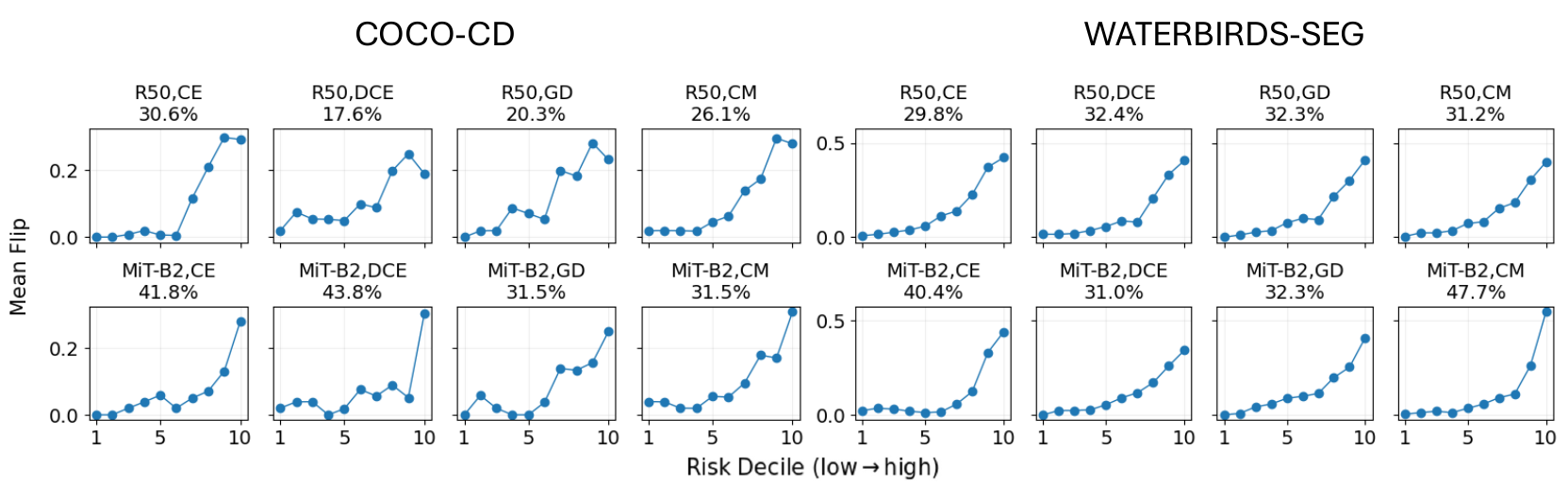}
\caption{\textbf{Mask-conditioned risk stratification.}
We compare decile plots obtained by computing risk over predicted-foreground pixels (deployable) versus ground-truth foreground pixels (analysis).
Both choices stratify within-foreground Flip similarly, suggesting the score reflects identity uncertainty rather than foreground support errors. The plots are obtained for $\rho=0.95$.}
\label{fig:risk_predfg_gtfg}
\end{figure}

\end{document}